\pdfoutput=1

\PassOptionsToPackage{table}{xcolor}
\IfFileExists{l4dc2026.cls}{%
  \documentclass{l4dc2026}
}{%
  \documentclass{l4dc2026}
}
\usepackage{times}
\usepackage{bm}
\usepackage{amssymb}
\usepackage{booktabs,array,multirow}
\usepackage{wrapfig}
\usepackage{needspace}
\RestyleAlgo{ruled}
\DontPrintSemicolon
\LinesNumbered
\usepackage{hyperref}
\hypersetup{hidelinks}
\setcitestyle{numbers,square,comma}
\usepackage[table]{xcolor}
\definecolor{supplementred}{RGB}{255,235,235}
\jmlryear{2027}
\makeatletter
\newtheorem{controlproblem}{Problem}
\@ifundefined{remark}{}{}
\@ifundefined{example}{}{}
\@ifundefined{proposition}{\newtheorem{proposition}{Proposition}}{}
\@ifundefined{proof}{\newenvironment{proof}{\par\noindent\textit{Proof.}\ }{\hfill$\square$\par\medskip}}{}
\makeatother

\newcommand{\R}{\mathbb R}

\newcommand{\T}{\mathbb T}
\newcommand{\E}{\mathbb E}
\newcommand{\dd}{\mathrm d}
\newcommand{\sg}{\operatorname{sg}}
\newcommand{\Law}{\operatorname{Law}}
\RequirePackage{fix-cm}

\title[Flow Matching for Sampled-Data Control of Angular Distributions]{Flow Matching for Sampled-Data Control of Angular Distributions}
\author{%
  \Name{Xin Mao}\textsuperscript{1,\textdagger},\enspace
  \Name{Anqi Dong}\textsuperscript{2,\textdagger,*},\enspace
  \Name{Hongteng Xu}\textsuperscript{3},\enspace
  \Name{Kenan Zhang}\textsuperscript{2}\\
  \addr \textsuperscript{1}University of North Carolina at Chapel Hill\\
  \addr \textsuperscript{2}\'Ecole polytechnique f\'ed\'erale de Lausanne (EPFL) and NCCR Automation\\
  \addr \textsuperscript{3}Renmin University of China\\
  \addr \textsuperscript{*}Corresponding author%
}

\AddToHook{cmd/normalsize/after}{%
  \setlength{\abovedisplayskip}{2pt}
  \setlength{\belowdisplayskip}{2pt}
  \setlength{\abovedisplayshortskip}{1pt}
  \setlength{\belowdisplayshortskip}{1pt}
}

\begin{document}
\maketitle
\begingroup
\renewcommand{\thefootnote}{\fnsymbol{footnote}}
\footnotetext[2]{Contributed equally to this work.}
\endgroup

\vspace{-0.5in}
\begin{abstract}
Synchronization and clustering in oscillator networks depend on the joint distribution of the phases. Flow Matching can steer this distribution by learning an angular velocity field from shortest arcs between paired configurations. Each arc is traversed at constant speed, yet the resulting field can vary along its own trajectories. In this article, we study how to control angular transport on the circle and product torus when each controller evaluation selects a rate to hold until the next update. Using the same local velocity samples, Flow Matching learns the instantaneous angular rate and MeanFlow learns its average along the reference trajectory over the control interval. Under direct angular actuation, holding the exact average gives the reference displacement. We relate the displacement error of holding the instantaneous rate and the training residual of MeanFlow to the conditional covariance of the teacher velocities. Experiments with 100 phases show that endpoint pairing affects the comparison. MeanFlow gives smaller one-step errors under minibatch pairing, while permutation alignment reverses the ranking for exchangeable phase clusters.
\end{abstract}


\section{Introduction}
A basic question in generative modeling is how to transport samples from one probability distribution to another. Flow Matching learns a velocity field for this transport by regressing on the velocities of paths joining paired samples~\citep{lipman2022flow}. In Euclidean space, a common choice is straight-line interpolation~\citep{liu2022flow}. Each path has constant velocity equal to its endpoint displacement, so a single Euler step is exact along that path. Regression averages these velocities at each state and time. The resulting field can vary along its own trajectories even when every training path is straight. This has led to methods that straighten learned trajectories for sampling with fewer evaluations~\citep{pooladian2023multisample,yue2025oat}.

In distributional control, the system dynamics determine which motions the inputs can produce~\citep{chen2016optimal,chen2021transport,chen2024density}. For a linear system, Flow Matching can use paths generated by feasible inputs and learn a feedback law from those inputs~\citep{mei2024flow}. When feedback is available only at discrete times, the controller must choose an input for the whole interval until its next update. The velocity along the desired trajectory may change during that interval. MeanFlow addresses the corresponding sampling problem by learning the average velocity between two times~\citep{geng2026mean}. In sampled-data swarm control, this idea has been used to learn a coefficient that determines a minimum-energy input waveform between updates~\citep{dong2026learning}.

A single angular phase lies on a circle, and a configuration of $n$ phases lies on the product torus $\T^n$. Straight interpolation between distinct points on a circle follows a chord that leaves the circle. Riemannian Flow Matching uses geodesics to construct paths on manifolds~\citep{chen2024flow}. Traversing a shortest arc at constant speed over a unit interval gives an angular velocity equal to its signed rotation. Molecular torsion angles also lie on a product of circles, where torsional diffusion can be used to sample molecular conformers~\citep{jing2022torsional}. PPFlow uses torsional flow matching to design peptides for a given protein target~\citep{lin2024ppflow}, and FlowPacker uses it to predict side-chain conformations given a protein sequence and backbone~\citep{lee2025flowpacker}. In oscillator networks, synchronization and clustering depend on the relations among phases~\citep{dorfler2014synchronization,zlotnik2016phase}. For example, independent uniform phases and perfectly synchronized phases with a uniform common angle have identical one-phase marginals. Their joint laws distinguish the two patterns.

In this article, we study how to steer a joint distribution of directly actuated phases with few controller evaluations. Each evaluation selects a rate for each phase to hold until the next update. Flow map matching and Riemannian MeanFlow describe the transport between two times~\citep{boffi2025flowmap,woo2026riemannian,zhong2026riemannian}. Direct actuation gives this interval transport a simple physical realization. Using the same shortest-arc teachers, we train Flow Matching to predict the instantaneous rate and MeanFlow to predict the average rate along the resulting flow until the next update. Holding the exact average rotates each phase to its reference endpoint.

The choice of endpoint pairs determines the reference flow for both controllers. We consider minibatch optimal transport across configurations~\citep{tong2023improving} and, for exchangeable phases, permutation alignment within a pair~\citep{klein2023equivariant}. For independent draws from exchangeable endpoint laws, we show that minimum-cost alignment with uniform tie breaking preserves the target law without increasing the cost of any sampled transfer. We relate the displacement error from holding the exact instantaneous rate to the divergence of the density-weighted covariance of the teacher velocities. The same covariance enters the training residual at the exact MeanFlow field through the flow Jacobian. In a clustering experiment with 100 phases and minibatch pairing, one evaluation gives mean circular $W_2$ errors of $0.180$ radians for MeanFlow and $0.444$ for Flow Matching. Permutation alignment reduces the Flow Matching error to $0.137$ radians and reverses the ranking. For an ordered twisted pattern, MeanFlow retains an advantage at low evaluation budgets.

\section{Preliminaries and control problem}\label{sec:problem}
\subsection{Conditional Flow Matching and interval averages}
Let $\rho_0$ and $\rho_1$ be probability laws on $\R^d$ with finite second moments. Conditional Flow Matching (CFM) constructs a path between these laws by mixing probability paths whose velocities are known~\citep{lipman2022flow,tong2023improving}. Let $p_t(\cdot\mid\xi)$ denote the path selected by $\xi$ and $b_t(\cdot\mid\xi)$ its velocity field. Draw $\xi$, an independent uniform time $t\in[0,1]$, and a state $x_t\sim p_t(\cdot\mid\xi)$. The corresponding teacher velocity is $w_t=b_t(x_t\mid\xi)$. Minimizing $\E\|v_\theta(x_t,t)-w_t\|^2$ over velocity fields gives the instantaneous velocity $v(x,t)=\E[w_t\mid x_t=x,t]$. For straight interpolation, $\xi=(x_0,x_1)$ is a pair drawn from a coupling $\pi$ of the endpoint laws. At time $t$ the state is $x_t=(1-t)x_0+tx_1$ and the teacher velocity is $w_t=x_1-x_0$. Optimal transport CFM (OT-CFM) selects the coupling by minimizing the expected squared displacement~\citep{tong2023improving}. Under a population-optimal coupling, the interpolated laws trace a Wasserstein geodesic~\citep{mccann1997convexity}. In the Benamou--Brenier formulation~\citep{benamou2000computational}, these laws and their regression field minimize the kinetic action subject to the continuity equation and the prescribed endpoints. The minimum is $\int_0^1\E\|v(x_t,t)\|^2\,\dd t=W_2^2(\rho_0,\rho_1)$. Computing the full coupling gives the optimum for the empirical laws. Minibatch methods reduce this cost by solving a smaller assignment between fresh batches of equal size~\citep{gabriel2019computational,pooladian2023multisample,tong2023improving}. MeanFlow uses the same local velocity samples to learn the average velocity along a trajectory of the regression field~\citep{geng2026mean}. Suppose a particle starts at $x$ at time $t$ and follows $v$ to $y$ at time $r>t$. The average velocity is $\bar v(x,t,r)=(y-x)/(r-t)$. Holding this rate gives the same displacement even if the particle changes velocity along the way.

\subsection{Angular control and sampling}\label{sec:angular-setup}
Given laws $\rho_0$ and $\rho_1$ on $\T^n=(\mathbb S^1)^n$, we seek a controller for $\dot\varphi=u$, $u\in\R^n$, that steers $\varphi(0)\sim\rho_0$ to $\Law(\varphi(1))=\rho_1$. Each sample is a configuration of $n$ phases, with $z_k=e^{\mathrm{i}\varphi_k}$. On a grid $0=t_0<\cdots<t_K=1$, the controller observes the configuration at $t_\ell$ and holds its input until $t_{\ell+1}$. At fixed $K$, we assess terminal accuracy and the expected effort per phase, $\mathcal E(u)=(2n)^{-1}\E\int_0^1\|u_t\|^2\,\dd t$. To sample a shortest-arc teacher on the flat torus~\citep{chen2024flow}, draw $(\varphi_0,\varphi_1)\sim\pi$ and an independent $t\sim\operatorname{Unif}(0,1)$. Let $\Delta_k=\measuredangle(z_{0,k},z_{1,k})\in[-\pi,\pi)$ be the signed shortest rotation, choosing $-\pi$ at antipodal points. The sampled state is $\varphi_t=(\varphi_0+t\Delta)\bmod 2\pi$ and its angular rate is $w_t=\Delta$. For the normalized distance $d_n(z_0,z_1)^2=\|\Delta\|^2/n$, this path has effort $d_n(z_0,z_1)^2/2$, the minimum for the pair. Minimizing its expectation over couplings gives $\tfrac12 W_{2,d_n}^2(\rho_0,\rho_1)$~\citep{villani2009optimal}. MeanFlow also samples an end time $r$, independently of the endpoints given $t$. With probability $d$ we set $r=t$ and otherwise draw $r$ uniformly from $(t,1)$.

\section{Learning angular transport under sampled-data control}\label{sec:complex-fm}
For an interval $[t,r]$ of length $h=r-t$, the controller can hold one of two rates. The instantaneous rate $v$ is the conditional mean of the sampled teacher rate $w_t$. The interval rate is the average $\bar v$ along the flow of $v$ over $[t,r]$.

\subsection{From paired configurations to instantaneous feedback}\label{sec:teacher}
The teacher of Section~\ref{sec:angular-setup} has lift $\widetilde\varphi_t=\widetilde\varphi_0+t\Delta$ from any representative $\widetilde\varphi_0\in\R^n$. In complex coordinates, $z_t=z_0\odot\exp(\mathrm{i}t\Delta)$ and $\dot z_t=\mathrm{i}z_t\odot\Delta$, with componentwise operations. Under direct angular actuation, the lifted rate $\Delta$ is the physical input. The choice of pairs determines which rates enter the regression. Minibatch optimal transport pairs whole configurations across two batches~\citep{tong2023improving}. When phases are interchangeable, symmetry also permits assignments within a pair~\citep{klein2023equivariant}. Two configurations that differ only in the order of their phases can, for example, be aligned at zero cost. Since the assignment depends on the sampled source, its effect on the target law requires justification. Let $P_\sigma$ denote a coordinate permutation and call a law exchangeable if it is invariant under every such permutation.
\begin{proposition}[Permutation coupling]\label{prop:permutation}
Draw $\varphi_0$ and $\varphi_1$ independently from exchangeable laws $\rho_0$ and $\rho_1$. Minimize $d_n(z_0,P_\sigma z_1)^2$ over permutations, choosing $\sigma^\star$ uniformly among minimizers. Then $(\varphi_0,P_{\sigma^\star}\varphi_1)$ has marginals $\rho_0,\rho_1$, and its cost is at most $d_n(z_0,z_1)^2$ for each sampled pair.
\end{proposition}
For the squared circular distance, sorting the phases and testing cyclic shifts solves this assignment~\citep{delon2010fast}. This gives a coupling for the exchangeable cluster task. In the twisted-pattern task the index of a phase specifies its position in the pattern, so we retain the coordinate labels. For a fixed coupling, let $\rho_t=\Law(\varphi_t)$ be the teacher marginal.
\begin{controlproblem}[Angular Flow Matching]\label{prob:fm}
Given samples $(\varphi_t,t,w_t)$ with teacher rate $w_t=\Delta$, learn the instantaneous angular rate $v_\theta(\varphi,t)$ by minimizing $\mathcal L_{\mathrm{FM}}(\theta)=n^{-1}\E\|v_\theta(\varphi_t,t)-w_t\|^2$ over periodic networks. The population target is
$v(\varphi,t)=\E[w_t\mid\varphi_t=\varphi,t].$
\end{controlproblem}
Assume $v$ generates a unique flow $\psi_{t,r}$ that carries $\rho_t$ to $\rho_r$. Under continuous feedback this field steers the initial law to the target. With sampled feedback, holding $v_\theta(\varphi,t)$ for $h=r-t$ produces the displacement $h v_\theta(\varphi,t)$. Even an exact instantaneous prediction can miss the reference endpoint when the velocity varies along the trajectory.

\subsection{Finite-interval control and angular effort}\label{sec:complex-mf}\label{sec:method}
MeanFlow describes transport over a finite interval by the average velocity along the flow~\citep{geng2026mean}. For sampled-data control we use forward intervals~\citep{dong2026learning}. On the torus, the terminal configuration specifies a rotation only up to integer multiples of $2\pi$. Starting from a real representative $\widetilde\varphi$, lift the reference trajectory continuously and denote its endpoint by $\widetilde\psi_{t,r}(\widetilde\varphi)$. The rate that reproduces its lifted displacement is
\begin{equation}\label{eq:average-velocity}
\bar v(\varphi,t,r)
=h^{-1}(\widetilde\psi_{t,r}(\widetilde\varphi)-\widetilde\varphi)
=h^{-1}\int_t^r v(\psi_{t,s}(\varphi),s)\,\dd s .
\end{equation}
Changing the initial representative translates both lifted endpoints by the same multiple of $2\pi$ in each coordinate. The rate is therefore well defined on $\T^n$. Under direct actuation it is also a physical input. Holding it for time $h$ rotates the phases to $\psi_{t,r}(\varphi)$, retaining the winding of the reference trajectory. The difference between holding $v$ and holding $\bar v$ depends on the acceleration of the reference flow. For straight training paths in Euclidean space, this acceleration is determined by the covariance of the teacher velocities conditional on the current state~\citep{tsimpos2025eulerian}. The lifted angular teachers are straight, which gives the following expression for the error over one update. All proofs are given in Appendix~\ref{app:angular-proofs}.
\begin{proposition}[Displacement error of a held angular rate]\label{prop:held-bias}
Let $p_s$ denote the density of $\rho_s$ with respect to flat volume on $\T^n$, and set $\Sigma_s(x)=\operatorname{Cov}(\Delta\mid\varphi_s=x)$ at fixed $s$. Suppose $p_s>0$ and $p_s$, $v$, and $\Sigma_s$ are continuously differentiable in state and time on $[t,r]$. Along $\gamma_s=\psi_{t,s}(\varphi)$, the lifted displacement error satisfies
\begin{equation*}
h\bigl(\bar v(\varphi,t,r)-v(\varphi,t)\bigr)
=-\int_t^r(r-s)\Big[\frac{\operatorname{div}(p_s\Sigma_s)}{p_s}\Big](\gamma_s)\,\dd s ,
\end{equation*}
where divergence acts on the rows of the matrix field.
\end{proposition}
For example, start with a uniform phase on the circle and independently choose a teacher rotation $+a$ or $-a$ with equal probability, $0<a<\pi$. At every time the two rates remain equally likely given the phase. The teacher variance is $a^2$, while the density and covariance are constant in space. Both exact controllers hold zero and reproduce the reference trajectory.

Using the exact interval rate also lets us compare update budgets at the same terminal law. For a grid $\mathcal T=\{t_0,\ldots,t_K\}$, let $\mathcal E_{\mathcal T}$ be the expected effort per phase when the controller holds $\bar v$ on each interval. The following result gives the cost of requiring the controller to pass through one additional reference state.
\begin{proposition}[Effort of exact interval control]\label{prop:grid-effort}
Suppose the reference flow transports the teacher marginals and has absolutely continuous trajectories with square-integrable angular rates. Start the exact interval controller from $\rho_0$. It reproduces $\rho_{t_\ell}$ at every update. Let $\mathcal T'$ be obtained by removing one interior update from $\mathcal T$. Along a common reference trajectory, denote the exact rates on the two adjacent intervals by $b_1,b_2$ and their lengths by $h_1,h_2$. Then
\begin{equation*}
\mathcal E_{\mathcal T}-\mathcal E_{\mathcal T'}
=\frac{h_1h_2}{2n(h_1+h_2)}\,\E\|b_1-b_2\|^2 .
\end{equation*}
On every grid, $\tfrac12 W_{2,d_n}^2(\rho_0,\rho_1)\leq\mathcal E_{\mathcal T}\leq(2n)^{-1}\E\|\Delta\|^2$. If the shortest-arc teacher coupling minimizes the expected squared torus distance, equality holds in both bounds.
\end{proposition}
Thus, removing an update preserves the reference states at the retained times and strictly reduces effort whenever the two adjacent rates differ with positive probability. For a learned field, changing the grid can also change the terminal law.

\subsection{Learning the interval rate and applying the control}
\begin{figure}[htbp]
\centering
\includegraphics[width=\linewidth]{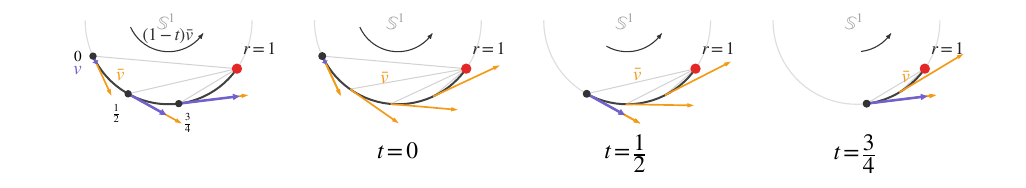}
\vspace{-0.3in}
\caption[Angular rates on the circle]{\small The left panel marks the three starting phases. The snapshots show $\mathrm{i}z_t v$ (blue) and $\mathrm{i}z_t\bar v$ (orange) on a common scale. Black arrows indicate the remaining rotations.}
\label{fig:angular-meanflow}
\vspace{-0.3in}
\end{figure}

When a phase speeds up along the circle, holding its instantaneous rate $v$ from time $t$ to $r=1$ leaves it short of the endpoint, as the three snapshots in Figure~\ref{fig:angular-meanflow} illustrate. To reach that endpoint with a constant input, the controller holds the average rate $\bar v$ over $[t,1]$, producing the full remaining rotation $(1-t)\bar v$. On $\T^n$ this rule applies to each phase, with each average rate approaching its instantaneous value as $t\uparrow1$.
Fix one reference trajectory $\dot\varphi_s=v(\varphi_s,s)$ and an arbitrary terminal time $r$. Following the construction of MeanFlow~\citep{geng2026mean}, multiply its average rate by the remaining time to obtain
\begin{equation*}
(r-t)\bar v(\varphi_t,t,r)=\int_t^r v(\varphi_s,s)\,\dd s .
\end{equation*}
Advance $t$ along this same trajectory. The original sampling formulation varies the upper integration limit. In forward control, $t$ is the lower limit, so the fundamental theorem of calculus gives $-v(\varphi_t,t)$ on the right. The product rule on the left yields
\begin{equation}\label{eq:displacement-product-rule}
-\bar v(\varphi_t,t,r)+(r-t)\frac{\dd}{\dd t}\bar v(\varphi_t,t,r)=-v(\varphi_t,t).
\end{equation}
The total derivative includes the change in the starting configuration. With $r$ fixed, the chain rule gives
$\frac{\dd}{\dd t}\bar v(\varphi_t,t,r)
=\partial_t\bar v(\varphi_t,t,r)+D_\varphi\bar v(\varphi_t,t,r)\,v(\varphi_t,t)$.
Substituting into~\eqref{eq:displacement-product-rule} and collecting terms gives the forward MeanFlow identity
\begin{equation}\label{eq:meanflow-identity}
\boxed{\bar v=v+(r-t)\bigl(\partial_t\bar v+D_\varphi\bar v\,v\bigr).}
\end{equation}
Conversely, this identity keeps the reconstructed endpoint constant along the trajectory. The boundary condition $h\bar v\to0$ as $t\uparrow r$ identifies it with the reference endpoint. For a continuously differentiable flow, the rate extends to the diagonal as $\bar v(\varphi,t,t)=v(\varphi,t)$.

\begin{controlproblem}[Angular MeanFlow]\label{prob:mf}
Given the teacher samples of Problem~\ref{prob:fm}, draw an end time $r\geq t$ independently of the endpoint pair conditional on $t$, as in Section~\ref{sec:angular-setup}. Learn the average angular rate~\eqref{eq:average-velocity} with a periodic network $\bar v_\theta(\varphi,t,r)$ using the loss $\mathcal L_{\mathrm{MF}}(\theta)=n^{-1}\E\|e_\theta\|^2$. The sample residual is
$e_\theta=\bar v_\theta-\sg[w_t+(r-t)\dot{\bar v}_\theta]$,
where $\sg(\cdot)$ stops the gradient through the target and $\dot{\bar v}_\theta=\partial_t\bar v_\theta+D_\varphi\bar v_\theta\,w_t$ differentiates along the sampled teacher path with $r$ fixed.
\end{controlproblem}
At $r=t$ the derivative term drops out and $\bar v_\theta(\varphi_t,t,t)$ is trained against $w_t$ by the Flow Matching loss of Problem~\ref{prob:fm}. For $r>t$, the stopped target is affine in $w_t$. Conditional on the state and the two times, its expectation therefore equals the target formed using $v(\varphi_t,t)$. Improved MeanFlow uses the average-velocity network to parameterize a regression on instantaneous velocity~\citep{geng2026improved}. For the stopped-target rule above, the effect of teacher variability depends on the sensitivity of the endpoint to its initial phase.
\begin{proposition}[Residual at the reference interval rate]\label{prop:mf-residual}
Suppose $\bar v$ and the lifted flow are continuously differentiable, with finite expectations below. Let $J=D\widetilde\psi_{t,r}(\widetilde\varphi_t)$. At $\bar v_\theta=\bar v$, the residual equals $e_\star=-J(w_t-v(\varphi_t,t))$ and has zero conditional mean given $(\varphi_t,t,r)$. Its mean-square value is
$\mathcal L_{\mathrm{MF}}\big|_{\bar v_\theta=\bar v}
=\frac1n\E\operatorname{tr}\bigl(J\Sigma_t(\varphi_t)J^\top\bigr)$.
For a parameterization realizing the exact field, stopped-gradient sample loss has zero expected parameter gradient whenever that gradient is integrable.
\end{proposition}
In the uniform-circle example, the reference flow is the identity and $J=1$. The exact interval field therefore has loss $a^2$ even though it reproduces every reference endpoint. The loss records the disagreement among local teacher samples.

At time $t_\ell$, Algorithm~\ref{alg:implementation} queries the MeanFlow network with the next update time $t_{\ell+1}$ and holds the predicted average rate for $h_\ell=t_{\ell+1}-t_\ell$. One can also query the same network on the diagonal, $\bar v_\theta(\boldsymbol\varphi_\ell,t_\ell,t_\ell)$, to estimate the instantaneous rate of Problem~\ref{prob:fm}. Only the query time changes. In both cases the input is held until $t_{\ell+1}$, giving
\begin{equation}\label{eq:sampled-update}
\boxed{\mathbf z_{\ell+1}=\mathbf z_\ell\odot\exp(\mathrm{i}h_\ell\mathbf u_\ell).}
\end{equation}
Each hold uses one network evaluation and incurs effort $h_\ell\|\mathbf u_\ell\|^2/(2n)$. The Flow Matching baseline in Section~\ref{sec:case-studies} uses its separately trained instantaneous field $v_\theta(\boldsymbol\varphi_\ell,t_\ell)$ for Euler execution.

\subsection{Known linear drift}\label{sec:linear-drift}
On a local lift, let $x=\widetilde\varphi-\widetilde\varphi^\star$ denote the phase deviation from a reference configuration, so that $z=\exp(\mathrm{i}(\widetilde\varphi^\star+x))$ componentwise. Suppose $\dot x=Ax+Bu$, with $(A,B)$ controllable and input weight $R=R^\top\succ0$. For Kuramoto oscillators with a common natural frequency and independent inputs, linearization near synchrony in a rotating frame gives $A=-L$ and $B=I$, where $L$ is the coupling Laplacian~\citep{kuramoto1984chemical,dorfler2014synchronization}. With $\Phi(r,t)=e^{A(r-t)}$, the weighted controllability Gramian is
$W(t,r)=\int_t^r\Phi(r,\tau)BR^{-1}B^\top\Phi(r,\tau)^\top\,\dd\tau$.
For paired lifted endpoints $(x_0,x_1)$, set $\lambda_{01}=W(0,1)^{-1}\bigl(x_1-\Phi(1,0)x_0\bigr)$. The minimum-energy teacher input is
$u_t=R^{-1}B^\top\Phi(1,t)^\top\lambda_{01}$.
At a sampled time $t$, the teacher state is $x_t=\Phi(t,0)x_0+W(0,t)\Phi(1,t)^\top\lambda_{01}$ and its velocity is $w_t=Ax_t+Bu_t$~\citep{chen2016optimal,mei2024flow}.
\Needspace{30\baselineskip}
\begin{wrapfigure}{R}{0.48\textwidth}
\vspace{-\intextsep}
\hfill\begin{minipage}{\dimexpr\linewidth-1em\relax}
\small
\SetAlCapFnt{\small\bfseries}
\SetAlCapNameFnt{\small\normalfont}
\SetInd{0.4em}{0.45em}
\begin{algorithm2e}[H]
\caption{Angular MeanFlow training}\label{alg:training}
\KwIn{Coupling $\pi$, diagonal fraction $d$.}
\For{each training batch}{
Draw endpoint pairs from $\pi$ and independent $t\sim\operatorname{Unif}(0,1)$\;
Compute the signed rotations $\Delta$\;
Set $\varphi_t\leftarrow(\varphi_0+t\Delta)\bmod 2\pi$\;
Draw $r\sim d\,\delta_t+(1-d)\operatorname{Unif}(t,1)$\;
$X\leftarrow(\varphi_t,t,r)$, $\eta\leftarrow(\Delta,1,0)$\;
$(b,\dot b)\leftarrow\operatorname{jvp}(\bar v_\theta,X,\eta)$\;
$e\leftarrow b-\sg[\Delta+(r-t)\dot b]$\;
Update $\theta$ using the batch mean of $\|e\|^2/n$\;
}
\end{algorithm2e}
\begin{algorithm2e}[H]
\caption{Angular MeanFlow control}\label{alg:implementation}
\KwIn{Trained $\bar v_\theta$, grid $\{t_\ell\}_{\ell=0}^K$.}
\For{$\ell=0,\ldots,K-1$}{
Measure $\varphi$ at $t=t_\ell$\;
Set $r\leftarrow t_{\ell+1}$\;
$\mathbf u_\ell\leftarrow\bar v_\theta(\varphi,t,r)$\;
Hold $\mathbf u_\ell$ until $t_{\ell+1}$\;
}
\end{algorithm2e}
\end{minipage}
\vspace{-\intextsep}
\end{wrapfigure}
The conditional input $\bar u(x,t)=\E[u_t\mid x_t=x,t]$ defines the reference velocity $v(x,t)=Ax+B\bar u(x,t)$ and its flow $\psi_{t,r}$. Following control-space MeanFlow~\citep{dong2026learning}, we learn an interval coefficient $c_\theta(x,t,r)$. A fixed coefficient $c$ generates the minimum-energy waveform $u(\tau)=R^{-1}B^\top\Phi(r,\tau)^\top c$ on $[t,r]$. Its endpoint is $\Phi(r,t)x+W(t,r)c$, so the exact coefficient satisfies
$c(x,t,r)=W(t,r)^{-1}\bigl(\psi_{t,r}(x)-\Phi(r,t)x\bigr)$.
Write $S=\Phi(r,t)BR^{-1}B^\top\Phi(r,t)^\top$. Differentiating the endpoint relation along the reference flow with $r$ fixed gives $Sc=W\dot c+\Phi B\bar u$,
where $\Phi=\Phi(r,t)$, $W=W(t,r)$, and $\dot c=\partial_t c+D_xc\,v$. Training uses the teacher input $u_t$ and the sampled derivative $\partial_t c_\theta+D_xc_\theta\,w_t$ on the right-hand side. At execution, one network evaluation fixes $c$ for the interval, while the waveform varies with $\tau$ through $\Phi(r,\tau)$. For $A=0$ and $B=R=I$, the Gramian becomes $hI$ and $S=I$. The coefficient is then the average angular rate $\bar v$, recovering~\eqref{eq:meanflow-identity} and the constant-rate update~\eqref{eq:sampled-update}. Numerical evaluation of the Laplacian model for trajectories that remain near synchrony is left to future work.

\section{Numerical Experiments}\label{sec:case-studies}
\begin{wraptable}{r}{0.55\textwidth}
\vspace{-0.28in}
\caption{Angular transport studies.}
\label{tab:angular-studies}
\small
\setlength{\tabcolsep}{3pt}
\begin{tabular}{@{}>{\raggedright\arraybackslash}p{0.35\linewidth}>{\raggedright\arraybackslash}p{0.23\linewidth}>{\raggedright\arraybackslash}p{\dimexpr0.42\linewidth-12pt\relax}@{}}
\toprule
Configuration & State space & Target \\
\midrule
100 phases & $\T^{100}$ & Phase clusters \\
100 phases on a ring & $\T^{100}$ & Twisted state \\
20 and 50 phases & $\T^{20}$, $\T^{50}$ & Phase clusters\\
\bottomrule
\end{tabular}
\vspace{-\intextsep}
\end{wraptable}
We compare Flow Matching and MeanFlow at small evaluation budgets on two tasks in $\T^{100}$, phase clustering and an ordered twisted state (Table~\ref{tab:angular-studies}). Both methods use the same shortest-arc teachers for each coupling. We use batch optimal transport on both tasks and test the permutation coupling of Proposition~\ref{prop:permutation} on the exchangeable cluster target. Appendix~\ref{app:scaling} reports the cluster results for 20 and 50 phases.

\subsection{Common setup}\label{sec:setup}
Initial phases are independent and uniform on the circle, and the horizon is $[0,1]$. Batch optimal transport (batch OT) pairs batches of 200 whole configurations by minimizing the total squared distance $d_n^2$ while retaining the phase labels. We train the flat-torus versions of Flow Matching (FM) and MeanFlow~\citep{chen2024flow,woo2026riemannian,zhong2026riemannian} with the same architecture and training budget. At each update, held-rate FM applies its predicted instantaneous rate and MeanFlow applies its predicted interval average. Integrated FM takes one Heun or fourth-order Runge--Kutta (RK4) step between updates, evaluating intermediate rates at predicted configurations. For $K\in\{1,2,4,8,16,32,64\}$ updates, the number of network evaluations (NFE) is $K$ for the held rates, $2K$ for Heun and $4K$ for RK4. Table~\ref{tab:phase-comparison} fixes $K=1$, and Figure~\ref{fig:nfe} compares terminal errors at equal NFE. We compute the effort of each held input exactly and report a numerical estimate for integrated FM.

We include three reference controllers. The stationary controller applies zero input. Phase feedback holds $u=\gamma\nabla_\varphi V$ with $V(\varphi)=\sum_k\log\nu(\varphi_k)$ for target phase density $\nu$ and selects $\gamma$ on validation data. Target sampling pairs test states with fresh target configurations by batch OT and follows the shortest arcs to those targets. All methods use the same 1024 initial test states, with observations only at grid times. Training, validation, and test data are disjoint. We report five training seeds for the 100-phase cluster task with batch OT and three for the other experiments. Distances are in radians, with independent target samples providing a reference for sampling variability.

\subsection{Phase clusters}\label{sec:clusters}
A configuration has empirical phase measure $\mu_z^n=n^{-1}\sum_{k=1}^n\delta_{z_k}$. The target density is the von Mises mixture
$\nu(\alpha)=\allowbreak\sum_{a=1}^{N_{\mathrm g}}w_a\frac{\exp(\kappa\cos(\alpha-\beta_a))}{2\pi I_0(\kappa)}$,
where $\alpha\in[-\pi,\pi)$ is a scalar angle, $I_0$ is the modified Bessel function, and the $N_{\mathrm g}$ positive weights sum to one. We use $\kappa=16$ and three groups at $(-2\pi/3,0,2\pi/3)$ with weights $(0.2,0.5,0.3)$. Sampling $\rho_1$ allocates exactly $nw_a$ phases per group, randomly permutes assignments, and draws component phases. Fixed counts induce dependence, without prescribing coordinatewise transport. The positive component widths give a distributional phase pattern rather than exact phase collapse.

\begin{figure}[t]
\centering
\includegraphics[width=\linewidth]{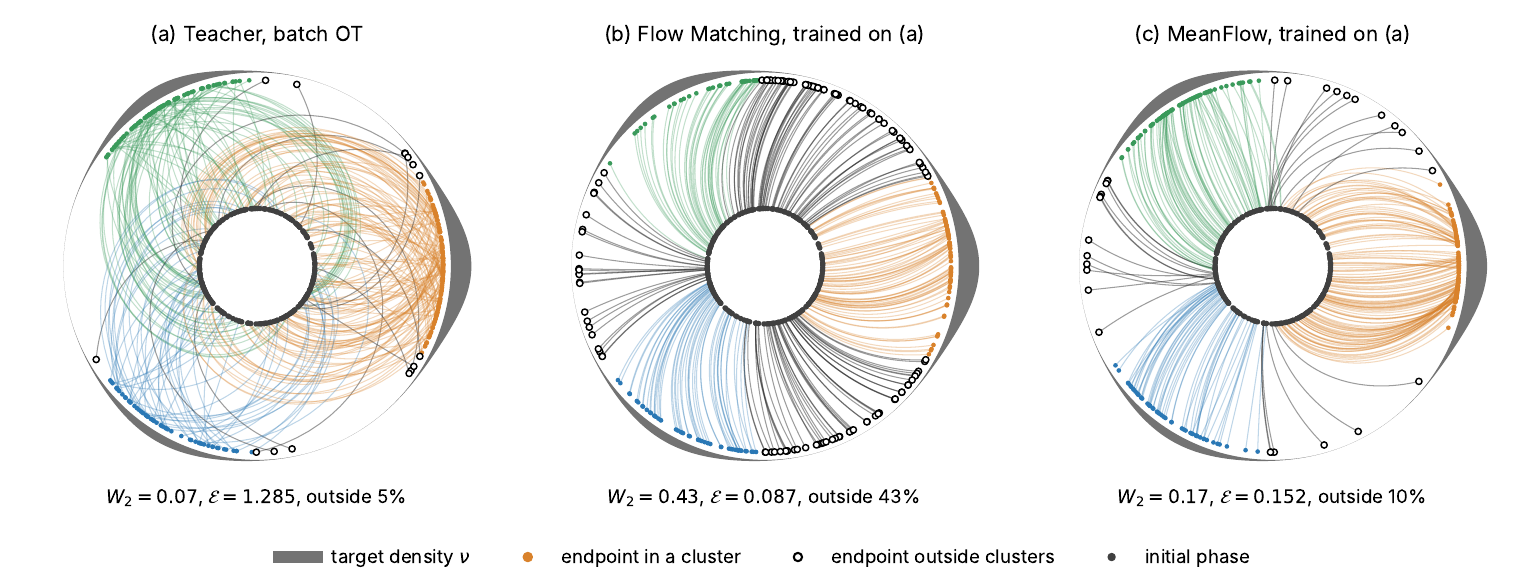}
\vspace{-0.3in}
\caption{\small Cluster trajectories at one update ($K=1$). (a) Batch-OT teachers. (b) Held-rate Flow Matching. (c) MeanFlow. Both controllers train on (a). Angle denotes phase and radius time, so constant rates trace spirals. The inner ring shows three initial test configurations and the gray band the target density $\nu$. Filled endpoints lie within two component standard deviations of a cluster, hollow endpoints outside. Circular $W_2$, effort $\mathcal E$, and the fraction outside summarize 1024 test configurations from one seed.}
\label{fig:trajectories}
\vspace{-0.3in}
\end{figure}

\paragraph{\noindent Couplings.}
For 100 phases, batch OT reduces the teacher effort per phase from $1.64$ for independent pairs to $1.29$. Figure~\ref{fig:trajectories}a shows crossings in the plotted trajectories of individual phases. Aligning the phases within each pair by Proposition~\ref{prop:permutation} reduces the teacher effort to $0.16$.

\paragraph{Metrics.}
The primary metric is the mean circular $W_{2,\mathbb S^1}(\mu_z^n,\nu_n)$ over configurations, using angular geodesic distance~\citep{delon2010fast}. The reference $\nu_n$ assigns mass $1/n$ to each of $n$ quantiles of the distribution with density $\nu$. In the full configuration space, empirical $W_{2,d_n}$ is $1.46$ between independent target samples and $1.56$ between initial and target samples. To compare the exchangeable laws, we therefore also report the Wasserstein distance between the laws of their empirical patterns $\mu_z^n$, using circular $W_2$ as the ground distance (pattern $W_2$). Permutation equivariance of the controllers preserves exchangeability of the initial law. Group occupancies measure how closely the terminal weights match the target.

\begin{table}[t]
\caption{\small Cluster task with batch OT at one update ($K=1$). Errors are means $\pm$ standard deviations over five seeds. NFE counts network evaluations. Bold entries give the lowest errors among learned methods. Independent target samples give a mean circular $W_2$ of $0.068$.}
\label{tab:phase-comparison}
\small
\begin{tabular}{@{}p{0.39\linewidth}cccc@{}}
\toprule
Method & Circle $W_2$ & Pattern $W_2$ & $\mathcal E$ & NFE \\
\midrule
Stationary & $0.563\pm0.004$ & $0.561\pm0.006$ & $0$ & $0$ \\
Phase feedback & $0.555\pm0.005$ & $0.555\pm0.006$ & $0.004$ & $0$ \\
Target sampling + shortest arc & $0.068\pm0.000$ & $0.063\pm0.000$ & $1.286$ & $0$ \\
Angular FM, held & $0.444\pm0.007$ & $0.440\pm0.009$ & $0.090$ & $1$ \\
Angular FM, integrated (Heun) & $0.531\pm0.017$ & $0.538\pm0.022$ & $0.085$ & $2$ \\
Angular FM, integrated (RK4) & $0.297\pm0.013$ & $0.292\pm0.014$ & $0.118$ & $4$ \\
\rowcolor{red!10}
Angular MeanFlow & $\mathbf{0.180}\pm0.006$ & $\mathbf{0.169}\pm0.008$ & $0.163$ & $1$ \\
\bottomrule
\end{tabular}
\end{table}

\begin{figure}[t]
\centering
\includegraphics[width=\linewidth]{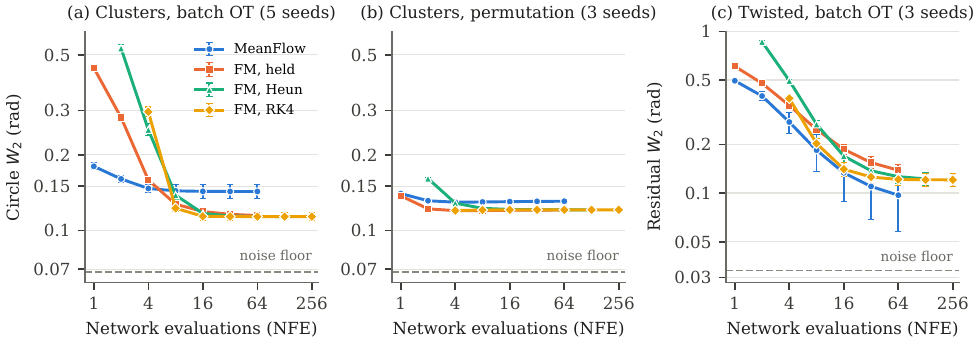}
\caption{\small Terminal error versus network evaluations (NFE), mean $\pm$ standard deviation over seeds. (a) Clusters with batch OT. (b) Clusters with permutation coupling. (c) Twisted state with batch OT. Errors are circular $W_2$ in (a,b) and residual $W_2$ in (c). Dashed lines mark the sampling-noise floor.}
\label{fig:nfe}
\vspace{-0.4in}
\end{figure}

\paragraph{Few evaluations.}
With batch OT, one MeanFlow evaluation gives a mean circular $W_2$ of $0.180$, compared with $0.444$ for held-rate FM (Table~\ref{tab:phase-comparison}). The corresponding group occupancies are $(0.20,0.49,0.30)$ and $(0.29,0.41,0.30)$. Many phases under held-rate FM stop short of the clusters (Figure~\ref{fig:trajectories}b). Proposition~\ref{prop:held-bias} gives the displacement error from holding the exact field. The learned rollouts can also incur error from estimating the field. MeanFlow gives lower circular $W_2$ in every seed at one and two evaluations and in four of five seeds at four evaluations ($0.147$ versus $0.159$ on average). At two and four evaluations, held-rate FM also outperforms a single Heun or RK4 step, respectively. Phase feedback gives group weights near $1/3$ and validation selects the smallest $\gamma$ tested. With the same hyperparameters at $n=20$ and $50$, MeanFlow gives smaller errors in every seed at one and two evaluations, while FM is comparable or better from eight evaluations onward.

\paragraph{Role of the coupling.}
Permutation alignment changes the one-evaluation circular $W_2$ from $0.444$ to $0.137$ for held-rate FM and from $0.180$ to $0.140$ for MeanFlow (Figure~\ref{fig:nfe}b). Held-rate FM then has a slightly lower mean error at every tested budget.

\paragraph{Many evaluations.}
With batch OT, FM overtakes MeanFlow at eight evaluations ($0.127$ for held-rate FM and $0.122$ for RK4, against $0.144$ for MeanFlow). At larger budgets MeanFlow levels off at $0.143$, while FM reaches $0.113$.

\subsection{Twisted state}\label{sec:twisted}
The second target is a noisy twisted state on a ring, $\varphi_k=\alpha+2\pi k/n+\epsilon_k$, where $\alpha$ is uniform and the perturbations $\epsilon_k$ are independent von Mises variables with concentration $\kappa=16$. With $\epsilon_k=0$, this is the $1$-twisted phase-locked state of a ring of coupled oscillators~\citep{wiley2006size}. Both $\rho_0$ and $\rho_1$ have uniform phase marginals, so the structure exists only at the configuration level: $\rho_1$ concentrates near a closed curve in $\T^{100}$, and transport must order the phases along the ring. The target is invariant under common rotations and cyclic shifts of the index. To preserve the ring order, we retain phase labels and use batch OT, with teacher effort $1.41$ per phase.

\paragraph{Metrics.}
For a configuration, let $\hat\alpha$ and $R$ be the phase and magnitude of $\frac1n\sum_{k} e^{\mathrm i(\varphi_k-2\pi k/n)}$, and define the residual pattern
$\mu_{\mathrm{res}}^n=\frac1n\sum_{k=1}^n\delta_{\exp(\mathrm i(\varphi_k-\hat\alpha-2\pi k/n))}$.
The residual $W_2$ is the mean of $W_{2,\mathbb S^1}(\mu_{\mathrm{res}}^n,\eta_n)$ over configurations, where $\eta_n$ places mass $1/n$ to each of $n$ quantiles of the centered von Mises distribution with concentration $\kappa$. Subtracting $\hat\alpha$ removes the uniform common rotation, so this distance measures whether the phases follow the ring order with the prescribed spread. The twisted order parameter $R$ equals one for an exact twist and $0.97$ for target samples, and we verify that $\hat\alpha$ remains uniform. The sampling-noise floor of the residual $W_2$ is $0.033$.

\paragraph{Few evaluations.}
Phase feedback uses only the phase density $\nu$, which is uniform here, so it reduces to zero input; the initial configuration has residual $W_2$ $1.47$ and $R=0.09$. MeanFlow is more accurate than held-rate FM at every budget in all seeds (Figure~\ref{fig:nfe}c): $0.495$ against $0.608$ with one evaluation, and $0.275$ against $0.348$ with four. It also outperforms the best integrated FM at equal evaluations up to four evaluations in all seeds. A single evaluation forms the twist only partially, with $R=0.79$.

\paragraph{Many evaluations.}
Unlike in the cluster task, MeanFlow continues to improve with more evaluations, reaching $0.097\pm0.039$ at $64$ evaluations against $0.121$ for RK4, and remains more accurate than integrated FM from eight evaluations onward in two of three seeds.

\paragraph{Cost and references.}
Training MeanFlow takes $1.7$--$1.9$ times as long as training FM in these experiments and requires a Jacobian--vector product. At execution, both learned policies dispense with online target sampling and pairing. Their reported effort per phase is at most $0.29$ for clusters and $1.1$ for the twisted state, below the corresponding batch-OT teacher efforts, with larger terminal errors than target sampling.

\section{Conclusion}\label{sec:discussion}
From common shortest-arc teachers on the torus, Flow Matching learns instantaneous angular rates and MeanFlow their interval averages. In 100-phase clustering, minibatch pairing favored MeanFlow at one update, while permutation alignment favored Flow Matching. For the ordered twisted state, where no alignment within configurations applies, MeanFlow remained more accurate at low budgets. Future work will address coupled dynamics, including Kuramoto phases, Stuart--Landau amplitudes and phases~\citep{nakao2016phase}, and Acrobot control through a single elbow torque~\citep{spong1995swing}. For molecular conformer generation, we will investigate whether a few learned torsional updates preserve the populations of low-energy conformers.

\bibliography{references}

\setcounter{theorem}{0}
\providecommand{\theHtheorem}{}
\renewcommand{\theHtheorem}{app.\arabic{theorem}}

\setlength{\emergencystretch}{2em}
\appendix
\section{Theoretical details}\label{app:angular-proofs}
The propositions are restated with the numbering used in the main text.

\subsection{Notation and reference flow}\label{app:notation}
A configuration of $n$ phases is a point $\varphi\in\T^n=(\mathbb S^1)^n$, with complex coordinates $z_k=e^{\mathrm i\varphi_k}$.
The phases are directly actuated by $\dot\varphi=u$ over $[0,1]$.
Given a coupling $\pi$ of the initial and target laws $\rho_0,\rho_1$, draw $(\varphi_0,\varphi_1)\sim\pi$ and define the signed shortest rotations
\[
\Delta_k=\measuredangle(z_{0,k},z_{1,k})\in[-\pi,\pi).
\]
At antipodal points we choose $-\pi$.
The teacher path and its constant angular rate are
\[
\varphi_s=(\varphi_0+s\Delta)\bmod 2\pi,
\qquad w_s=\Delta.
\]
The normalized squared torus distance is $d_n(z_0,z_1)^2=\|\Delta\|^2/n$.
We also write $d_n$ for the corresponding distance in phase coordinates, and $W_{2,d_n}$ for its quadratic Wasserstein distance.
A law is exchangeable if it is invariant under every coordinate permutation $P_\sigma$.

Let $\rho_s=\Law(\varphi_s)$, and let $p_s$ denote its density with respect to flat volume when this density exists.
The instantaneous regression field and the conditional teacher covariance are
\[
v(x,s)=\E[\Delta\mid\varphi_s=x],
\]
\[
\Sigma_s(x)=\operatorname{Cov}(\Delta\mid\varphi_s=x).
\]
We assume, whenever the reference flow is used, that $\dot\gamma_s=v(\gamma_s,s)$ generates a unique flow $\psi_{t,r}$ carrying $\rho_t$ to $\rho_r$.
Starting from a representative $\widetilde\varphi\in\R^n$, lift each trajectory continuously and denote its endpoint by $\widetilde\psi_{t,r}(\widetilde\varphi)$.
For $h=r-t>0$, the interval rate is
\[
\bar v(\varphi,t,r)
=\frac{\widetilde\psi_{t,r}(\widetilde\varphi)-\widetilde\varphi}{h}.
\]
Its value is independent of the chosen representative.
For a continuously differentiable flow, the rate extends to the diagonal by $\bar v(\varphi,t,t)=v(\varphi,t)$.

On a grid $\mathcal T=\{0=t_0<\cdots<t_K=1\}$, the exact interval controller holds $\bar v(\varphi,t_\ell,t_{\ell+1})$ until the next update.
Its expected effort per phase is
\[
\mathcal E_{\mathcal T}
=\frac{1}{2n}\E\int_0^1\|u_s\|^2\,\dd s.
\]
For MeanFlow training, draw $t$ independently of the endpoints and draw $r\geq t$ independently of the endpoints conditional on $t$.
At $X=(\varphi_t,t,r)$, the sampled residual is
\[
e_\theta
=\bar v_\theta(X)
-\sg\!\left[w_t+(r-t)
\left(\partial_t\bar v_\theta(X)+D_\varphi\bar v_\theta(X)\,w_t\right)\right],
\]
where $\sg$ holds the target fixed during differentiation.
The loss is $\mathcal L_{\mathrm{MF}}(\theta)=n^{-1}\E\|e_\theta\|^2$.
The proof below uses the forward identity
\[
\bar v=v+(r-t)(\partial_t\bar v+D_\varphi\bar v\,v),
\]
with the terminal time $r$ fixed.

\subsection{Permutation coupling (Proposition 1)}
\label{app:proof-1}

\begin{proposition}[Permutation coupling]\label{supp:prop:permutation}
Draw $\varphi_0$ and $\varphi_1$ independently from exchangeable laws $\rho_0$ and $\rho_1$. Minimize $d_n(z_0,P_\sigma z_1)^2$ over permutations, choosing $\sigma^\star$ uniformly among minimizers. Then $(\varphi_0,P_{\sigma^\star}\varphi_1)$ has marginals $\rho_0,\rho_1$, and its cost is at most $d_n(z_0,z_1)^2$ for each sampled pair.
\end{proposition}

\begin{proof}
Let $X$ and $Y$ be independent draws from the exchangeable laws $\rho_0$ and $\rho_1$. For fixed $x,y$, let $\mathcal A(x,y)$ be the set of coordinate permutations $P$ minimizing $d_n(x,Py)^2$. Draw $P^\star$ uniformly from $\mathcal A(X,Y)$ and set $\widetilde Y=P^\star Y$. The set of permutations is finite and each cost is continuous, so this rule defines a measurable conditional law.

Fix a coordinate permutation $S$. Since $d_n(Sx,SPy)=d_n(x,Py)$, left multiplication by $S$ maps $\mathcal A(x,y)$ bijectively onto $\mathcal A(Sx,y)$. The uniform tie rule therefore relabels the aligned target by $S$ in law, including when several minimizers give the same configuration. Independence and exchangeability of the source give $(SX,Y)\overset{d}{=}(X,Y)$. It follows that $S\widetilde Y\overset{d}{=}\widetilde Y$.

For a bounded measurable function $b$, let $b_{\rm sym}(y)$ be its average over all coordinate permutations of $y$. On each draw, $\widetilde Y$ and $Y$ have the same coordinates up to permutation, so $b_{\rm sym}(\widetilde Y)=b_{\rm sym}(Y)$. By exchangeability,
\[
\E b(\widetilde Y)=\E b_{\rm sym}(\widetilde Y)
                  =\E b_{\rm sym}(Y)=\E b(Y).
\]
Thus $(X,\widetilde Y)$ has the prescribed marginals. Taking the identity permutation as a competitor gives $d_n(X,\widetilde Y)^2\leq d_n(X,Y)^2$ for every pair.
\end{proof}

\subsection{Displacement error of a held angular rate (Proposition 2)}
\label{app:proof-2}

\begin{proposition}[Displacement error of a held angular rate]\label{supp:prop:held-bias}
Let $p_s$ denote the density of $\rho_s$ with respect to flat volume on $\T^n$, and set $\Sigma_s(x)=\operatorname{Cov}(\Delta\mid\varphi_s=x)$ at fixed $s$. Suppose $p_s>0$ and $p_s$, $v$, and $\Sigma_s$ are continuously differentiable in state and time on $[t,r]$. Along $\gamma_s=\psi_{t,s}(\varphi)$, the lifted displacement error satisfies
\begin{equation*}
h\bigl(\bar v(\varphi,t,r)-v(\varphi,t)\bigr)
=-\int_t^r(r-s)\left[\frac{\operatorname{div}(p_s\Sigma_s)}{p_s}\right](\gamma_s)\,\dd s ,
\end{equation*}
where divergence acts on the rows of the matrix field.
\end{proposition}

\begin{proof}
Each lifted teacher path has constant rate $\Delta$. For a smooth periodic test function $\chi$, differentiation along the lift is valid even when a phase crosses the coordinate cut. Since $\|\Delta\|\leq\pi\sqrt n$ and $\nabla\chi$ is bounded, we may also differentiate under the expectation. Taking $a=1$ or $a=\Delta_k$ gives
\[
\frac{\dd}{\dd s}\E[a\,\chi(\varphi_s)]
=\E[a\,\nabla\chi(\varphi_s)\cdot\Delta].
\]
Write $p_s$ for the density of $\varphi_s$ and $\Sigma_s(x)=\operatorname{Cov}(\Delta\mid\varphi_s=x)$. For $a=1$, conditioning on $\varphi_s$ and integrating by parts on the torus gives the continuity equation $\partial_s p_s+\operatorname{div}(p_s v)=0$. For $a=\Delta_k$, the same calculation uses the $k$th row of the conditional second moment $vv^\top+\Sigma_s$. Collecting the rows gives
\[
\partial_s(p_s v)+\operatorname{div}\bigl(p_s(vv^\top+\Sigma_s)\bigr)=0,
\]
where divergence acts on each row. The assumed differentiability makes both equations pointwise. Subtracting $v$ times the continuity equation and dividing by $p_s>0$ gives the acceleration along a continuous lift of $\gamma_s=\psi_{t,s}(\varphi)$,
\[
\ddot{\widetilde\gamma}_s
=(\partial_s v+D_\varphi v\,v)(\gamma_s,s)
=-\left[\frac{\operatorname{div}(p_s\Sigma_s)}{p_s}\right](\gamma_s).
\]

Over $h=r-t$, holding the initial rate gives displacement $hv(\varphi,t)$. The reference displacement is $h\bar v(\varphi,t,r)=\widetilde\gamma_r-\widetilde\gamma_t$. Integrating the change in velocity and reversing the order of integration yields
\[
\begin{aligned}
h\bigl(\bar v(\varphi,t,r)-v(\varphi,t)\bigr)
&=\int_t^r\!\int_t^\tau\ddot{\widetilde\gamma}_s\,\dd s\,\dd\tau=-\int_t^r(r-s)\left[\frac{\operatorname{div}(p_s\Sigma_s)}{p_s}\right](\gamma_s)\,\dd s .
\end{aligned}
\]
In particular, if the divergence vanishes along the reference trajectory, holding its initial rate recovers its lifted endpoint.
\end{proof}

\subsection{Effort of exact interval control (Proposition 3)}
\label{app:proof-3}

\begin{proposition}[Effort of exact interval control]\label{supp:prop:grid-effort}
Suppose the reference flow transports the teacher marginals and has absolutely continuous trajectories with square-integrable angular rates. Start the exact interval controller from $\rho_0$. It reproduces $\rho_{t_\ell}$ at every update. Let $\mathcal T'$ be obtained by removing one interior update from $\mathcal T$. Along a common reference trajectory, denote the exact rates on the two adjacent intervals by $b_1,b_2$ and their lengths by $h_1,h_2$. Then
\begin{equation*}
\mathcal E_{\mathcal T}-\mathcal E_{\mathcal T'}
=\frac{h_1h_2}{2n(h_1+h_2)}\,\E\|b_1-b_2\|^2 .
\end{equation*}
On every grid, $\tfrac12 W_{2,d_n}^2(\rho_0,\rho_1)\leq\mathcal E_{\mathcal T}\leq(2n)^{-1}\E\|\Delta\|^2$. If the shortest-arc teacher coupling minimizes the expected squared torus distance, equality holds in both bounds.
\end{proposition}

\begin{proof}
Draw $\Gamma_0\sim\rho_0$ and follow the reference trajectory $\Gamma_s=\psi_{0,s}(\Gamma_0)$ with angular rate $V_s=v(\Gamma_s,s)$. On a grid interval $I_\ell=[t_\ell,t_{\ell+1}]$ of length $h_\ell$, the exact interval rate is $b_\ell=h_\ell^{-1}\int_{I_\ell}V_s\,\dd s$. Holding $b_\ell$ from $\Gamma_{t_\ell}$ gives the same lifted displacement as the reference and therefore reaches $\Gamma_{t_{\ell+1}}$. Starting from the same initial draw, the two trajectories agree at every update by induction. Since the reference flow transports the teacher marginals, the controller has law $\rho_{t_\ell}$ at each update.

Removing one update merges two adjacent intervals. Their lifted displacements add, so the merged rate is $b=(h_1b_1+h_2b_2)/(h_1+h_2)$. The effort on all other intervals is the same. Expanding the squared rates gives
\[
h_1\|b_1\|^2+h_2\|b_2\|^2-(h_1+h_2)\|b\|^2
=\frac{h_1h_2}{h_1+h_2}\|b_1-b_2\|^2 .
\]
Taking expectations and dividing by $2n$ gives the change in effort per phase. It is strictly positive precisely when the two rates differ with positive probability.

To compare with the teacher effort, first condition on the phase at a fixed time $s$. The teacher and reference states have the same law, and the conditional mean teacher rate is $v(\varphi_s,s)$. The conditional variance identity gives
\[
\E\|\Delta\|^2=\E\|V_s\|^2+\E\operatorname{tr}\Sigma_s(\varphi_s).
\]
Next, on each reference trajectory, $V_s-b_\ell$ integrates to zero over $I_\ell$. Expanding $\|V_s\|^2$ about $b_\ell$ on each interval and integrating the variance identity over the unit horizon yields
\[
\begin{aligned}
\E\|\Delta\|^2-2n\mathcal E_{\mathcal T}
&=\int_0^1\E\operatorname{tr}\Sigma_s(\varphi_s)\,\dd s+\sum_\ell\E\int_{I_\ell}\|V_s-b_\ell\|^2\,\dd s .
\end{aligned}
\]
Both terms are nonnegative, giving the upper bound.

For the lower bound, the executed endpoints form a coupling of $\rho_0$ and $\rho_1$. The integrated input $\int_0^1u_s\,\dd s$ gives one lifted displacement between them. Their squared torus distance is at most the squared norm of this displacement divided by $n$. Taking expectations and applying Cauchy--Schwarz on the unit interval gives
\[
W_{2,d_n}^2(\rho_0,\rho_1)
\leq\frac1n\E\left\|\int_0^1u_s\,\dd s\right\|^2
\leq 2\mathcal E_{\mathcal T}.
\]
For an optimal shortest-arc teacher coupling, $\E\|\Delta\|^2/n=W_{2,d_n}^2(\rho_0,\rho_1)$, so the two bounds coincide.
\end{proof}

\subsection{Residual at the reference interval rate (Proposition 4)}
\label{app:proof-4}

\begin{proposition}[Residual at the reference interval rate]\label{supp:prop:mf-residual}
Suppose $\bar v$ and the lifted flow are continuously differentiable, with finite expectations below. Let $J=D\widetilde\psi_{t,r}(\widetilde\varphi_t)$. At $\bar v_\theta=\bar v$, the residual equals $e_\star=-J(w_t-v(\varphi_t,t))$ and has zero conditional mean given $(\varphi_t,t,r)$. Its mean-square value is
\begin{equation*}
\mathcal L_{\mathrm{MF}}\big|_{\bar v_\theta=\bar v}
=\frac1n\E\operatorname{tr}\bigl(J\Sigma_t(\varphi_t)J^\top\bigr).
\end{equation*}
For a parameterization realizing the exact field, the stopped-gradient sample loss has zero expected parameter gradient whenever that gradient is integrable.
\end{proposition}

\begin{proof}
Condition on $X=(\varphi_t,t,r)$. Given $t$, the end time $r$ is sampled independently of the endpoints. The conditional mean of $w_t$ given $X$ is therefore $v=v(\varphi_t,t)$, and its conditional covariance is $\Sigma_t(\varphi_t)$.

Suppose the network realizes the differentiable interval field $\bar v$. For $r>t$, the reference identity gives $\bar v=v+h(\partial_t\bar v+D_\varphi\bar v\,v)$, where $h=r-t$ and all derivatives are evaluated at $X$. The sampled target uses $w_t$ in place of $v$. Subtracting it from the prediction gives
\[
\begin{aligned}
e_\star
=\bar v-w_t-h(\partial_t\bar v+D_\varphi\bar v\,w_t)
=-(I+hD_\varphi\bar v)(w_t-v).
\end{aligned}
\]
To identify the matrix in this expression, differentiate the lifted endpoint $\widetilde\varphi_t+h\bar v(X)$ with respect to its starting phase. Its Jacobian is $J=I+hD_\varphi\bar v$. On the diagonal $r=t$, the endpoint map is the identity and $\bar v(\varphi_t,t,t)=v$. Thus $e_\star=-J(w_t-v)$ also holds there.

Conditional on $X$, the matrix $J$ is fixed. The residual therefore has mean zero and covariance $J\Sigma_t(\varphi_t)J^\top$. Taking the trace and averaging over $X$ yields
\[
\frac1n\E\|e_\star\|^2
=\frac1n\E\operatorname{tr}\bigl(J\Sigma_t(\varphi_t)J^\top\bigr).
\]
When the target is held fixed during differentiation, the sample gradient is $(2/n)(D_\theta\bar v_\theta(X))^\top e_\star$. The network derivative is determined by $X$, so the conditional mean of the gradient is zero. Under the stated integrability assumption its expectation is zero as well.
\end{proof}

\section{Experimental details}\label{app:experiments}

Initial phases are independent and uniform.
The cluster target has weights $(w_1,w_2,w_3)=(0.2,0.5,0.3)$, centers $(-2\pi/3,0,2\pi/3)$, and von Mises concentration $\kappa=16$.
Each configuration has exactly $nw_a$ phases per group with randomly permuted assignments.
NFE counts network evaluations; MeanFlow and held-rate Flow Matching use one per update.

\subsection{Number of phases}\label{app:scaling}
Table~\ref{tab:scaling} repeats the cluster task with batch OT for $n=20$ and $n=50$, using the hyperparameters selected for $n=100$ and three seeds. Group counts $nw_a$ are integers for all three sizes. The noise floor grows as $n$ decreases, because each pattern contains fewer phases. For every size, MeanFlow is better than held-rate FM in every seed at one and two evaluations. One FM step stops short of the clusters at every size, with errors between $0.44$ and $0.47$. MeanFlow improves with $n$, from $0.294$ at $n=20$ to $0.180$ at $n=100$, partly because the noise floor decreases. From eight evaluations, FM is better in all seeds for $n=50$ and $n=100$, and in two of three for $n=20$.

\begin{table}[!ht]
\caption{Cluster task with batch OT for different numbers of phases: circular $W_2$ (rad), mean $\pm$ standard deviation over seeds, at equal NFE. Bold with a light red background: better in every seed. Hyperparameters are those selected for $n=100$.}
\label{tab:scaling}
\centering
\small
\begin{tabular}{cclcccc}
\toprule
$n$ & Floor & Method & NFE $=1$ & $2$ & $4$ & $8$ \\
\midrule
\multirow{2}{*}{$20$} & \multirow{2}{*}{$0.132$} & MeanFlow & \cellcolor{supplementred}$\mathbf{0.294}\pm0.001$ & \cellcolor{supplementred}$\mathbf{0.247}\pm0.003$ & $0.218\pm0.008$ & $0.203\pm0.011$ \\
 &  & FM, held & $0.471\pm0.010$ & $0.286\pm0.000$ & $0.224\pm0.000$ & $0.196\pm0.001$ \\
\midrule
\multirow{2}{*}{$50$} & \multirow{2}{*}{$0.092$} & MeanFlow & \cellcolor{supplementred}$\mathbf{0.229}\pm0.006$ & \cellcolor{supplementred}$\mathbf{0.202}\pm0.004$ & \cellcolor{supplementred}$\mathbf{0.176}\pm0.002$ & $0.168\pm0.004$ \\
 &  & FM, held & $0.438\pm0.006$ & $0.278\pm0.004$ & $0.186\pm0.002$ & \cellcolor{supplementred}$\mathbf{0.160}\pm0.002$ \\
\midrule
\multirow{2}{*}{$100$} & \multirow{2}{*}{$0.068$} & MeanFlow & \cellcolor{supplementred}$\mathbf{0.180}\pm0.006$ & \cellcolor{supplementred}$\mathbf{0.160}\pm0.005$ & $0.147\pm0.006$ & $0.144\pm0.008$ \\
 &  & FM, held & $0.444\pm0.007$ & $0.281\pm0.007$ & $0.159\pm0.005$ & \cellcolor{supplementred}$\mathbf{0.127}\pm0.004$ \\
\bottomrule
\end{tabular}
\end{table}

\subsection{Implementation details}\label{app:details}
For the cluster task, both networks use a shared per-phase perceptron.
Its inputs are the periodic features $(\sin p\varphi_k,\cos p\varphi_k)_{p\le3}$, the time inputs, and a mean-pooled embedding of all phases.
The resulting networks are permutation equivariant. For the twisted task, dilated circular convolutions over the phase index make them equivariant to cyclic shifts. Training uses Adam for $6000$ steps, with width $256$, batch $200$, and learning rate $10^{-3}$ for the cluster task, and width $64$, batch $100$, and learning rate $5\times10^{-4}$ for the twisted task. A fraction $d$ of the MeanFlow samples uses $r=t$. On validation data, $d\in\{0.25,0.5,0.75\}$ was selected by the mean error over $K\le4$, giving $d=0.5$ for the cluster task. The twisted task uses $d=0.25$. Learned models receive periodic state features and have equal validation budgets. Hyperparameters were selected with a separate seed on validation states. The phase-feedback gain $\gamma$ is selected on validation data from a logarithmic grid subject to $\gamma\kappa/K\le1$, and its achieved group weights and widths are measured rather than assumed.

\end{document}